\documentclass[letterpaper, 10 pt, conference]{ieeeconf}  %
\IEEEoverridecommandlockouts                              %

\usepackage{graphicx}
\usepackage{amsmath,amssymb,amsfonts}
\usepackage{svg}
\usepackage{multirow}

\usepackage{color}
\usepackage[normalem]{ulem} 
\definecolor{visnecurugu}{rgb}{0.6,0.1,0.6}

\title{\LARGE \bf
Vehicle Detection and Classification without Residual Calculation: Accelerating HEVC Image Decoding with Random Perturbation Injection
}

\author{Muhammet Sebul Berato\u{g}lu$^{1}$, Behçet U\u{g}ur T{\"{o}}reyin$^{1}$ 
\thanks{$^{1}$Signal Processing for Computational Intelligence Research Group, Informatics Institute, İstanbul Technical University, Maslak, Turkey 
      {\tt\small {\{beratoglu, toreyin\}@itu.edu.tr}}%
      }
\thanks{This work was supported by The Scientific and Technical Research Council of Turkey (TUBITAK) under the grant number 121E378.}
}

\begin{document}
\maketitle
\thispagestyle{empty}
\pagestyle{empty}

\begin{abstract}

In the field of video analytics, particularly traffic surveillance, there is a growing need for efficient and effective methods for processing and understanding video data. Traditional full video decoding techniques can be computationally intensive and time-consuming, leading researchers to explore alternative approaches in the compressed domain. This study introduces a novel random perturbation-based compressed domain method for reconstructing images from High Efficiency Video Coding (HEVC) bitstreams, specifically designed for traffic surveillance applications. To the best of our knowledge, our method is the first to propose substituting random perturbations for residual values, thereby creating a condensed representation of the original image while retaining information relevant to video understanding tasks, particularly focusing on vehicle detection and classification as key use cases. 

By not using any residual data, our proposed method significantly reduces the amount of data needed in the image reconstruction process, allowing for more efficient storage and transmission of information. This is particularly important when considering the vast amount of video data involved in surveillance applications. Applied to the public BIT-Vehicle dataset, we demonstrate a significant increase in the reconstruction speed compared to the traditional full decoding approach,  with our proposed random perturbation-based method being approximately 56\% faster than the pixel domain method. Additionally, we achieve a detection accuracy of 99.9\%, on par with the pixel domain method, and a classification accuracy of 96.84\%, only 0.98\% lower than the pixel domain method. Furthermore, we showcase the significant reduction in data size, leading to more efficient storage and transmission. Our research establishes the potential of compressed domain methods in traffic surveillance applications, where speed and data size are critical factors. The study's findings can be extended to other object detection tasks, such as pedestrian detection, and future work may investigate the integration of compressed and pixel domain information, as well as the extension of these methods to the full video decoding process, encompassing both intra and inter encoded bitstreams.

\end{abstract}

\section{Introduction}
\label{sec:introduction}
In recent years, the rapid growth of video data and the increasing demand for efficient video analytics have led researchers to seek new methods for analyzing and processing video streams. Traditional video processing techniques require decoding the entire video bitstream, which can be computationally intensive and time-consuming. In traffic surveillance applications, where real-time video understanding is crucial, the computational overhead associated with full video decoding can be a significant bottleneck. Consequently, researchers have begun to explore alternative approaches that leverage the compressed domain to reduce processing time and maintain the effectiveness of video understanding tasks \cite{comp1}, \cite{comp2}, \cite{comp3}. By directly analyzing the compressed video bitstream, it is possible to extract relevant information for video understanding tasks without the need for full decoding. 

In this study, we introduce a novel method for reconstructing images from HEVC bitstream by injecting random perturbations as a substitute for residual values, significantly speeding up the reconstruction process compared to standard intra decoding. To the best of our knowledge, our method is the first to propose substituting residual values instead of calculating them, thereby creating a condensed representation of the original image while retaining information pertinent to video understanding tasks, particularly focusing on vehicle detection and classification as key use cases. By operating directly in the compressed domain, our method avoids the computational overhead associated with full video decoding, leading to a more efficient and effective solution for traffic monitoring and management.

In video compression standards like HEVC, the encoded bitstream holds a substantial amount of data related to prediction error for image samples, known as residuals. For I-frames, these residuals make up about 85-90\% of the total data in the bitstream\cite{compressed_data_stream_HEVC}. Furthermore, residual coding accounts for an average of 77\% and 84\% of the total bits for dynamic continuous and discrete video textures, respectively\cite{rate_distortion_texture}. By not using any residual data, our proposed method significantly reduces the amount of data needed in the image reconstruction process, allowing for more efficient storage and transmission of information. This is particularly important when considering the vast amount of video data involved in surveillance applications.

To evaluate the performance of our proposed method, we conduct experiments on the public BIT-Vehicle dataset, a large-scale dataset comprising diverse vehicle types and imaging conditions. Our results demonstrate that the proposed method is able to reconstruct images approximately 56\% faster than the pixel domain method, while maintaining a high level of detection and classification accuracy. In particular, our method achieves a detection accuracy of 99.9\%, on par with the pixel domain method, and a classification accuracy of 96.84\%, only 0.98\% lower than the pixel domain approach. These results highlight the potential of our compressed domain method for traffic surveillance applications, where both speed and data size are critical factors.

Additionally, the increasing emphasis on data privacy and the need to comply with regulations such as the EU General Data Protection Regulation (GDPR) and the California Privacy Rights Act (CPRA) make data minimization a crucial consideration in video analytics. These regulations mandate that only the necessary data be collected to fulfill a certain purpose. A recent study presented a method to reduce the amount of personal data needed for machine learning predictions by removing or generalizing some input features of the runtime data, using knowledge distillation approaches \cite{data_minimization}. In the context of our proposed method, operating directly in the compressed domain and significantly reducing the amount of data needed for image reconstruction addresses the data minimization requirement set out in these regulations. By minimizing the data used for video understanding tasks, our method not only enhances processing efficiency but also helps organizations comply with privacy regulations.

By showcasing the potential of compressed domain methods for video understanding tasks in traffic surveillance applications, this study contributes to the growing body of research aimed at overcoming performance bottlenecks in video analytics. The faster reconstruction time and reduced data size associated with these methods make them a promising option for certain types of applications where speed and data size are important considerations. Further research into the potential applications and improvements of these methods could lead to significant advancements in the field of video analytics.

The remainder of this paper is organized as follows: Section 2 provides an overview of related work in object detection and video compression. Section 3 discusses the basic blocks of HEVC encoding and decoding, with a focus on intra-prediction. Section 4 presents our proposed approach for object detection and classification in compressed domain videos. Section 5 provides experimental results, and Section 6 concludes the paper.

\section{Related Works}
In this section, we review existing literature on vehicle classification in both the pixel and compressed domains, with a focus on methods using the BIT Vehicle dataset. We briefly discuss various techniques and recent advancements while highlighting how our approach differs from existing methods.
\subsection{Vehicle Classification in Pixel Domain}
Vehicle classification in the pixel domain has been a popular research topic over the years, with numerous review papers providing comprehensive overviews of various techniques and methods. Some notable review papers on this topic include those by Yang and Pun-Cheng \cite{pixel.7}, Wang et al. \cite{pixel.8}, and Zou et al. \cite{pixel.9}, which discuss the state-of-the-art in vehicle classification, as well as the challenges and future research directions.

In this section, we focus on methods that utilize the BIT Vehicle dataset, as this dataset allows us to directly compare our results with these works. We provide an overview of five representative vehicle classification methods based on the BIT Vehicle dataset.

Dong et al. \cite{pixel.10} proposed a vehicle type classification method using a semi-supervised convolutional neural network from vehicle frontal-view images. They introduced sparse Laplacian filter learning to obtain the filters of the network with large amounts of unlabeled data and trained the network on the challenging BIT-Vehicle dataset. The method demonstrated the effectiveness of using deep learning for vehicle classification in complex scenes.

Roecker et al. \cite{pixel.12} proposed a convolutional neural network model for vehicle type classification using low-resolution images from a frontal perspective. They trained the model on a subset of the BIT-Vehicle dataset and achieved an accuracy of 93.90\%, proving the model to be discriminative and capable of generalizing the patterns of the vehicle type classification task.

Sang et al. \cite{pixel.13} proposed a new vehicle detection model called YOLOv2\_Vehicle based on YOLOv2. They used the k-means++ clustering algorithm to cluster vehicle bounding boxes on the training dataset, improved the loss calculation method for bounding box dimensions, and adopted a multi-layer feature fusion strategy. The model achieved a mean Average Precision (mAP) of 94.78\% on the BIT-Vehicle validation dataset.

Wu et al. \cite{pixel.14} proposed a multi-scale vehicle detection method by improving YOLOv2 to address the foreground-background class imbalance and varying vehicle sizes in a scene. They introduced a new anchor box generation method called Rk-means++ and incorporated Focal Loss into YOLOv2 for vehicle detection. The method demonstrated better performance on vehicle localization and recognition on the BIT-Vehicle public dataset compared to other existing methods.

Taheri Tajar et al. \cite{pixel.11} developed a lightweight real-time vehicle detection model based on the Tiny-YOLOv3 network. They pruned and simplified the network and trained it on the BIT Vehicle dataset, achieving an mAP of 95.05\% and a detection speed of 17 fps, which is about two times faster than the original Tiny-YOLOv3 network.

In our work, we adopt the YOLOv7 framework \cite{yolov7} as the basis for our vehicle classification method. We focus on achieving comparable accuracy to pixel domain methods while operating in the compressed domain. By utilizing the strengths of YOLOv7 and adapting it to work with HEVC intra features, we aim to develop a computationally efficient vehicle classification method that maintains high accuracy.

\subsection{Related Works in Compressed Domain}
Recent years have witnessed a growing interest in developing object detection and classification methods in the compressed domain. In this section, we review some of the most relevant works and discuss how our approach differs from them.

Donghai Zhai et al. \cite{compressed.2.1} provided a comprehensive overview of object detection methods in the compressed domain across various video compression standards, including MPEG-2, H.264, and HEVC. They highlighted different ways of utilizing motion vector information for object detection and analyzed the techniques under various compression standards. Among the many works presented in their review, we have chosen the ones that focus on the HEVC compressed domain for a more detailed comparison with our approach.

Zhao et al. \cite{compressed.2.2} proposed a real-time moving object segmentation and classification method for surveillance videos using HEVC compressed domain features. Their approach only classified objects as persons or vehicles, while our method classifies vehicles into six specific types.

Chen et al. \cite{compressed.2.3} introduced a fast object detection method in the HEVC intra compressed domain. Their method used partitioning depths, prediction modes, and residuals for object detection, whereas our approach omits residuals and achieves good results with less computational demand.

Feng et al. \cite{compressed.2.4} proposed a fast framework for semantic video segmentation, named TapLab, which utilized motion vectors and residuals from compressed videos. Unlike their method, we focus on intra features and do not rely on motion vectors.

Choi and Bajic \cite{compressed.2.5} presented a human detection method based on HEVC intra coding syntax elements, including block size, intra prediction modes, and transform coefficient levels. Their approach did not require full bitstream decoding but focused on human detection rather than vehicle classification.

Wang et al. \cite{compressed.2.6} developed a highway vehicle counting method in the compressed domain using low-level features extracted from coding-related metadata. Their method is competitive with pixel-domain approaches in terms of computational cost but focuses on counting vehicles rather than classifying them into distinct types.

Our method differs from these works in several ways. We classify vehicles into six specific types, providing a more detailed classification for intelligent transportation applications. Furthermore, we are the first to suggest using random perturbation to reconstruct a frame without employing residual data, which makes our method less computationally demanding. While most of these works rely on motion vectors, our approach exploits the potential of intra features. Since a video consists of intra and inter frames, incorporating motion vectors in future works could further enhance our method. By using the state-of-the-art YOLOv7, we demonstrate close accuracy to pixel domain methods, showcasing the effectiveness of our approach.

\section{High Efficiency Video Coding (HEVC)}
The High Efficiency Video Coding (HEVC) is a video compression standard that was developed jointly by the ITU-T Video Coding Experts Group and the ISO/IEC Moving Picture Experts Group. It is designed to achieve higher compression efficiency compared to its predecessor, the H.264/MPEG-4 AVC standard. HEVC achieves higher compression efficiency by introducing new tools and techniques such as larger block sizes, more prediction modes, and more efficient entropy coding \cite{h264, h265}.

The HEVC compression algorithm processes the video data using a hierarchical organization of blocks. The encoding process begins with partitioning a frame into Coding Tree Units (CTUs), which are further partitioned into Coding Units (CUs), Transform Units (TUs), and Prediction Units (PUs). The prediction process can be either inter or intra. Intra prediction, also known as intra-frame prediction, is a technique to remove spatial correlation. It uses information from previously coded blocks within the same frame to predict the content of the current block. On the other hand, inter-prediction is used to remove the temporal correlation. It uses information from previously coded frames to predict the content of the current frame \cite{intra}. Our method is applied to intra predicted frames.

In intra prediction, the prediction is done by using reference samples and prediction modes. Reference samples are blocks of image or video data that are used as a reference for predicting the values of other blocks. They are extracted at the boundary from the upper and left blocks adjacent to the current PU. When reference samples are not available, they can be generated by copying samples from the closest available references. If no reference samples are available, a nominal average sample value (typically 128) is used in their place. HEVC uses several intra-prediction modes, such as Angular, DC, and Planar, to achieve better compression performance. The intra prediction modes use the same set of reference samples, and there are 33 prediction modes in total \cite{intra}.

The HEVC standard employs the discrete cosine transform (DCT) and the discrete sine transform (DST) to encode TUs. The residuals are the difference between predictions and original pixel values, and they are held in TUs. The block structure, prediction modes, and the quantized data are entropy coded and transmitted. 

\section{Compressed Domain Vehicle Detection and Classification}
Our proposed method for vehicle detection and classification in the compressed domain consists of two primary components: Image Reconstruction and Vehicle Detection and Classification. The Image Reconstruction component reconstructs an image based on the intra-prediction process in HEVC, eliminating the need for residual data. The Vehicle Detection and Classification component leverages the reconstructed image as input and employs the state-of-the-art YOLOv7 \cite{yolov7} to detect and classify vehicles.

This section delves into the details of each component, emphasizing their collaborative effort to achieve efficient and accurate vehicle detection and classification within the HEVC compressed domain.

\subsection{Image Reconstruction using Random Perturbations}
In this subsection, we discuss the reconstruction of images based on prediction unit information without resorting to residuals, consequently reducing data and computational requirements.

\begin{figure*}[t!]
\centering
    \includegraphics[width=\linewidth]{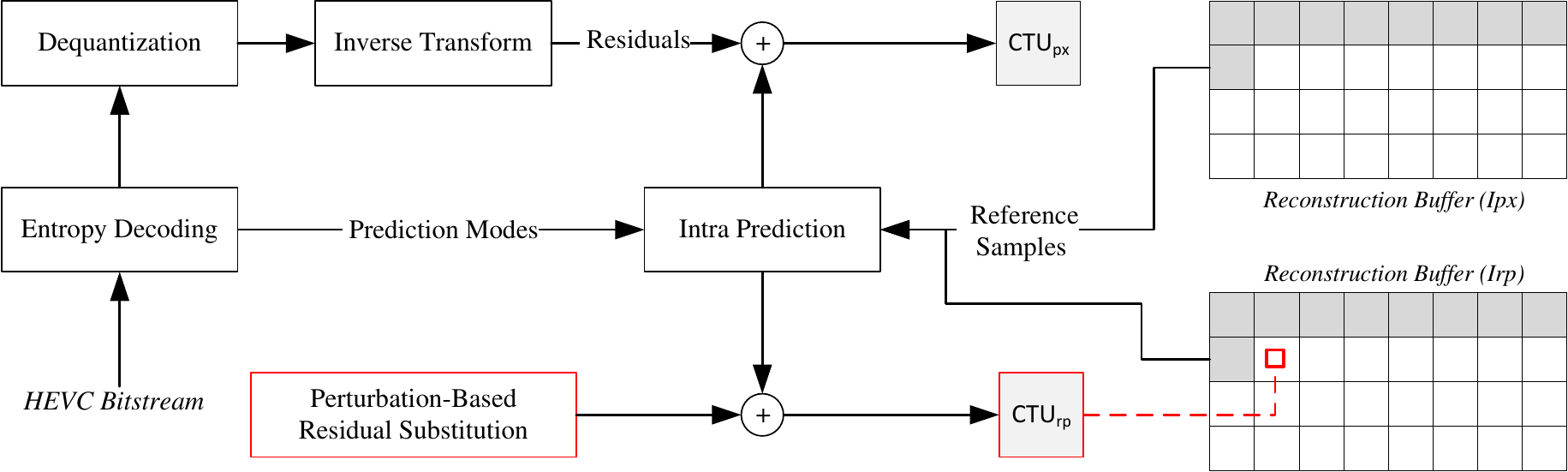}
    \caption{Reconstructing CTUs with HEVC Standard Decoding Process and with Random Perturbation Based Residual Substitution.}
    \label{fig04_5Methods}
\end{figure*}

Fig.\ref{fig04_5Methods} illustrates the process of reconstructing a coding tree unit (CTU) within the context of High Efficiency Video Coding(HEVC). The HEVC bitstream is first decoded by the entropy decoding block, which extracts syntax elements such as partition structure, prediction modes, and residual data. The decoder generates the CTU prediction by employing prediction modes and utilizing reference samples. The reference samples consist of neighboring pixels found within the image. The residual data is then added to the predicted CTU to generate the final $CTU_{px}$. Note that, the process of reconstructing CTUs with HEVC standard decoding process and estimated residuals is actually done in Coding Unit (CU) level. However, for simplicity and ease of understanding, we have presented the process in the context of CTUs.

\subsubsection{Standard Reconstruction}
Let $CU (x, y)$ be the intensity value of a reconstructed Coding Unit of an image $I$ at the spatial coordinates $(x, y)$. The intensity value of the CU can be expressed as the sum of its prediction value, $P (x, y)$, and its residual value, $R(x, y)$, as shown in the following equation:

\begin{equation}
CU (x, y) = P (x, y) + R(x, y)
\end{equation}

Decoding is an iterative process where the reconstructed CU serves as the input for subsequent CUs \cite{h265}. The prediction information's close relationship with residual information makes bypassing residuals a significant challenge \cite{compressed.2.5}. To address this, we explored potential signals that could replace residuals while maintaining the overall integrity of the reconstructed image.

\subsubsection{Impact of Ignoring Residuals}
In the case where $R(x, y)$ is assumed to be equal to 0, let's examine the consequences for the first CU to be decoded, which is situated at the top-left corner of the image. Due to the absence of reference pixels and in accordance with the HEVC standard, the reference value is assigned as 128 \cite{intra}. This value corresponds to the mean of the pixel range, equating to 128 for an 8-bit image. Consequently, the first reconstructed CU will be entirely gray. The first CU contains the reference pixels for subsequent CUs to be decoded. This means that for the second CU, the reference pixels will also be 128. As this process continues, all reference pixels will have a value of 128, leading to a fully gray predicted image, as depicted in (2):

\begin{equation}
I(x, y) = 128 \quad \text{if} \quad R(x, y) = 0
\end{equation}

\subsubsection{Impact of Constant Residuals}
In the case where $R(x, y)$ is assumed to be equal to a constant value, this constant would be added to all decoded CUs throughout the decoding process. As the process progresses, this addition would accumulate, causing saturation in the reconstructed image. The constant value of $R(x, y)$ would directly impact the resulting saturation level and might lead to a loss of detail or visual information in the reconstructed image. This illustrates the importance of properly handling the residual values to achieve accurate and high-quality image reconstructions.

\subsubsection{Random Perturbations as a Substitute for Residuals}

Taking these cases into account, we propose using a series of integers $Rp$, as a replacement for the residuals. These integers are based on a Gaussian distribution and can take both negative and positive values. The choice of using a Gaussian distribution to generate the estimated perturbations ($Rp$) is motivated by the fact that the Gaussian distribution is a commonly used model for representing the distribution of errors and noise in natural images. By modeling the estimated perturbations using a Gaussian distribution, we can generate a more realistic approximation of the actual residual values, which in turn results in a more accurate reconstruction of the CTUs.

The Gaussian distribution is given by the following formula:

\begin{equation}
f(x) = \frac{1}{\sqrt{2 \pi \sigma^2}} \exp\left(-\frac{(x - \mu)^2}{2 \sigma^2}\right)
\end{equation}

To generate the series $Rp$, we sample $n$ discrete integer random numbers from this distribution with varying standard deviations, such as $std = 1, 2, 3, 4, 5$:

\begin{equation}
Rp_n = {rp_1, rp_2, \dots, rp_n}
\end{equation}

Each $rp_i$ in the series $Rp$ is obtained by iteratively sampling from the Gaussian distribution until the sample set meets the desired mean ($\mu = 0$) and standard deviation ($\sigma = std$):

\begin{equation}
rp_i = \text{round}(f(x_i))
\end{equation}

where $x_i$ is a random sample from the Gaussian distribution, and the round function is used to obtain integer values.

Given the series $Rp$, we can now construct the predicted CTUs without residual data. For each pixel in a CTU, we replace the residual value with the corresponding value from the series $Rp$:

\begin{equation}
CU_{rp}(x, y) = P(x, y) + rp_i
\end{equation}

As we process the pixels in the CU, we use the next value from the series $Rp$, $rp_i$, as a replacement for the residual. The length of $Rp$ is determined by the maximum size of a CU, which is $64\times64 = 4096$. The series $Rp_n$, with a mean of zero and varying standard deviations, is generated once and used for all predicted image generations. This ensures that the predicted images are consistent and reproducible.The resulting reconstructed image using the random perturbation method will be referred to as $I_{rp}$.

\begin{figure}[t!]
\centering
    \includegraphics[width=\columnwidth]{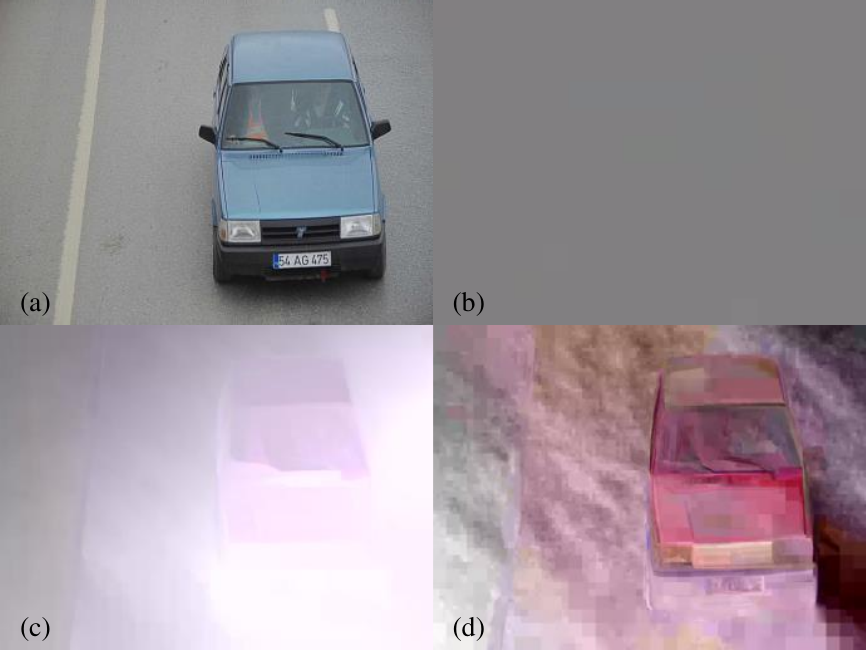}
\caption[Comparison of different approaches to substitute residuals.]{Comparison of different approaches to substitute residuals.(a) Standard reconstruction using HEVC. (b) Impact of Ignoring Residuals, $R(x, y) = 0$.   (c) Impact of Constant Residual, $R(x, y) = 1$. (d) Random Perturbations as a Substitute for Residuals, $I_{rp}$ with $\mu = 0$ and $\sigma = 7$.}
\label{fig03_4Figures}
\end{figure}

\begin{figure*}[t!]
\centering
    \includegraphics[width=\linewidth]{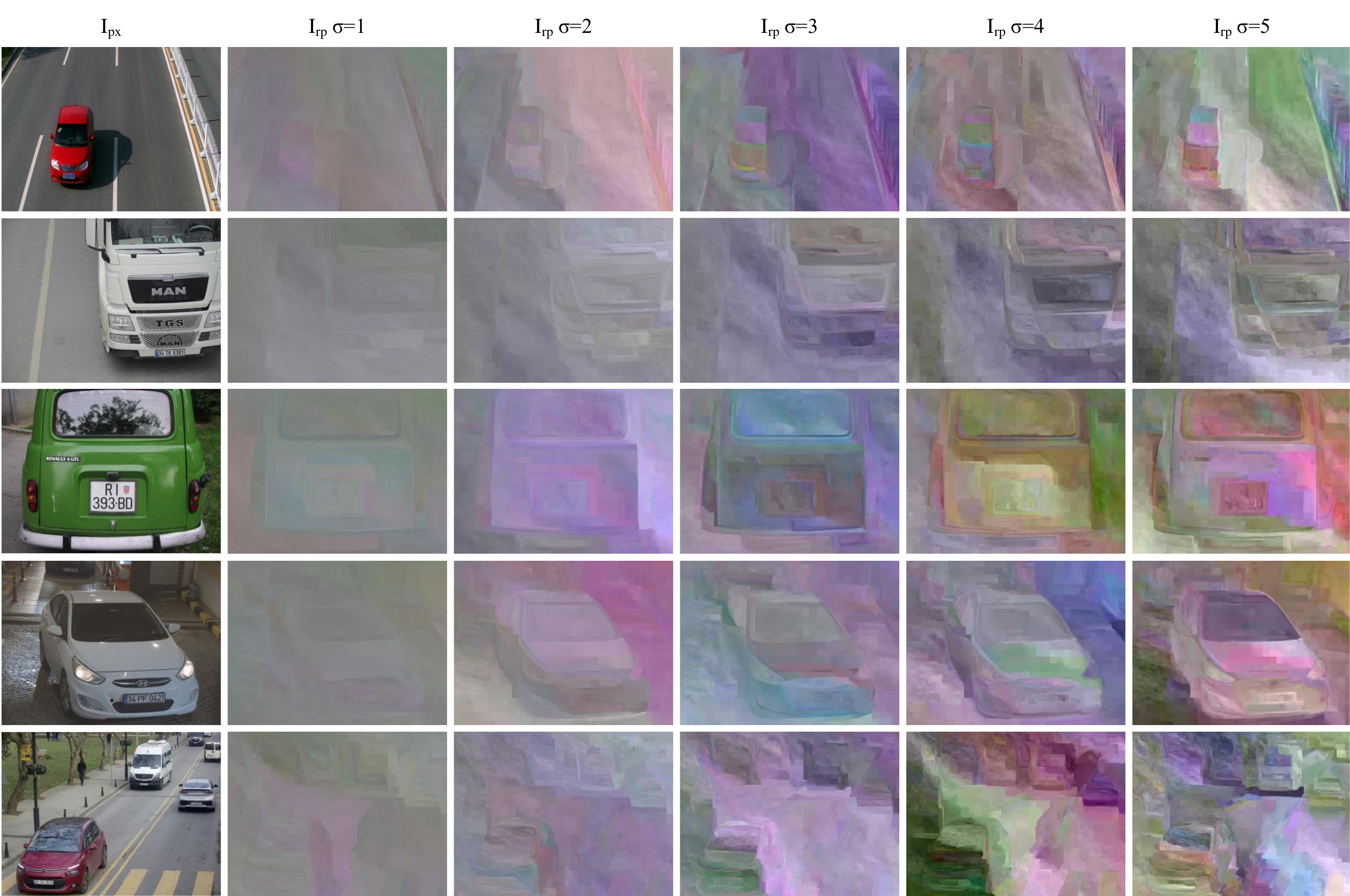}
\caption{Effect of Varying Standard Deviations in Random Perturbation for Image Reconstruction.}
 \label{fig4_MultipleReconstructions}
\end{figure*}

An example of the different approaches to substitute residuals is shown in Figure \ref{fig03_4Figures}. This figure provides a visual comparison of the original pixel domain image reconstructed using the standard HEVC method (a) and the effects of various approaches to substitute residuals on the image quality, such as ignoring residuals (b), using constant residuals (c), and employing random perturbations (d). The random perturbations approach, with a mean ($\mu$) of 0 and standard deviation ($\sigma$) of 7, clearly demonstrates a better representation of the original image compared to other approaches.

In Figure \ref{fig4_MultipleReconstructions}, we present a matrix of images showcasing the effectiveness of our Gaussian-based Random Perturbations method in reconstructing images of vehicles from various scenes while measuring the effect of different standard deviations. The first column displays pixel domain images, while the other columns represent reconstructed images using our method with different standard deviations. As illustrated, our method successfully constructs a close silhouette of the pixel domain image, retaining the general boundaries of objects in the frame and the necessary information for image classification. As the standard deviation increases, the image becomes more apparent to the human eye, yet classification can still be done successfully for all different standard deviations. Our approach bypasses the calculation of residual data, yet still enables the creation of an image with general boundaries retained, which serves as valuable input for object detection tasks.

\subsection{Vehicle Detection and Classification}

For the vehicle detection and classification task, we employ the state-of-the-art YOLOv7 object detector \cite{yolov7}. YOLOv7 is a single-stage, real-time detector that has demonstrated impressive speed and accuracy in real-time object detection tasks. According to its paper, YOLOv7 outperforms previous models and has set a significant benchmark in the field.

The YOLOv7 family includes the YOLOv7-Tiny model, which is the smallest in the family with just over 6 million parameters. Despite its compact size, the YOLOv7-Tiny model achieves a validation AP of 35.2\%, surpassing the performance of previous YOLO-Tiny models.

In our proposed method, we utilize the reconstructed images, generated using Random Perturbations, as input for the Vehicle Detection and Classification component. These reconstructed images, referred to as $I_{rp}$ images, are used to train a model using the Darknet framework \cite{darknet}. The trained model is then applied to the task of vehicle detection and classification within the HEVC compressed domain.

By leveraging the YOLOv7 object detector and the reconstructed images, our proposed method is able to efficiently and accurately detect and classify vehicles in the compressed domain, without the need for residual data.

\section{Experimental Results}

This section presents the experimental results of our proposed method. We begin with an overview of the experimental setup, detailing the hardware and software configurations used in the experiments, and dataset description. Next, we provide a comparison of time efficiency and accuracy with other relevant approaches. The results demonstrate the effectiveness of our method in achieving efficient and accurate vehicle detection and classification within the HEVC compressed domain.

\begin{figure*}[t!]
\centering
    \includegraphics[width=\linewidth]{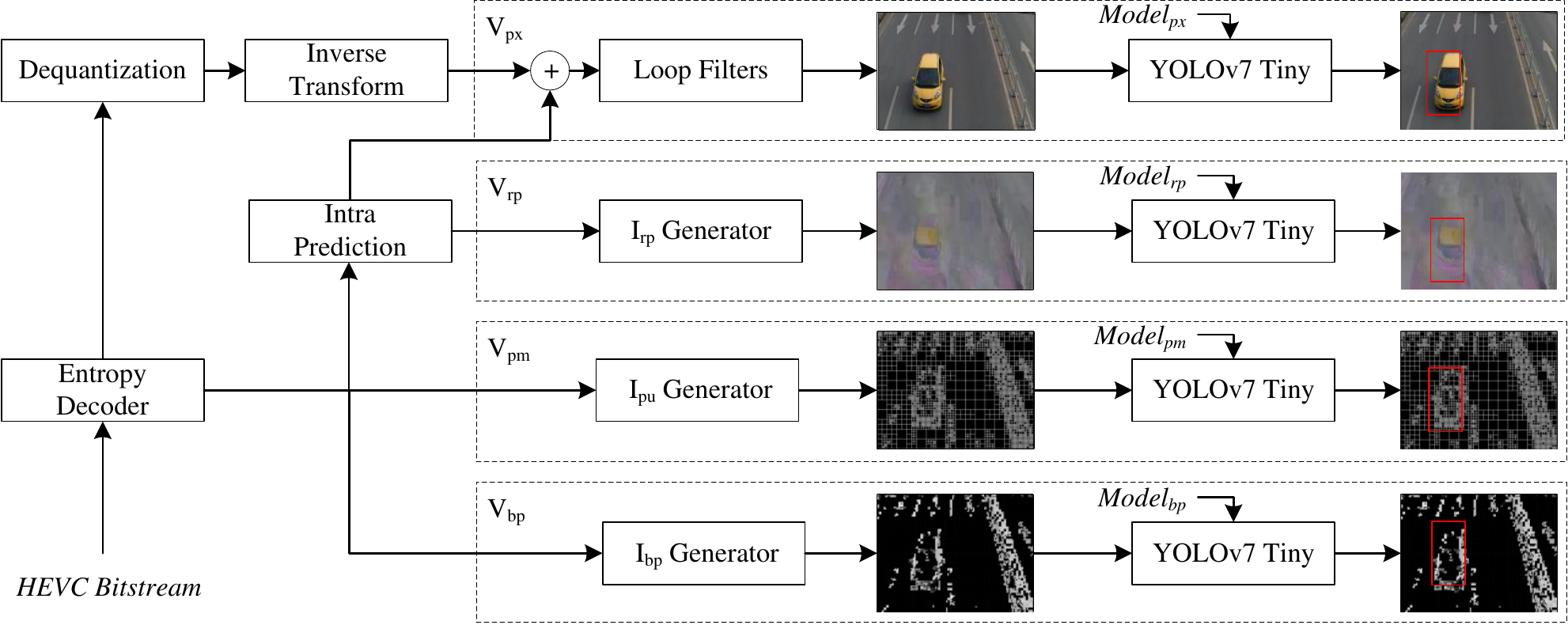}
    \caption{Comparison of four different input images generated from the same HEVC bitstream and fed into their corresponding YOLOv7-Tiny networks for vehicle detection and classification.}
    \label{fig6.1_ThesisVehicleMainFigures}
\end{figure*}

\subsection{The Experimental Setup}
The experimental setup for comparing our proposed method begins with obtaining HEVC bitstreams from the JPEG format images, $I_{org}$, in the BIT database through intra-encoding using the HEVC encoder \cite{h265.3}.

Then, from these bitstreams, we generate the pixel domain image $I_{px}$ and the Random Perturbation Images $I_{rp}$. Note that $I_{px}$ images are first encoded and then decoded from $I_{org}$ to ensure a fair comparison, as each source image faces the same HEVC encoding distortion. To further compare with previous compressed domain approaches \cite{beratoglu01, beratoglu02}, we also generate $I_{bp}$ (block partition based) and $I_{pu}$ (prediction unit based) images from the same bitstream.

Next, separate YOLOv7-Tiny models are trained for each method under the following conditions: the same number of images with the same resolution are used for training and testing; the same deep learning network structure and hyperparameters are utilized, including "learning rate", "batch size", "number of epochs to train for", and "number of nodes in the given layer". A total of 250,000 batches are conducted to ensure that the average loss no longer decreases, and the weight with the best Mean Average Precision (mAP) is chosen from among the generated weights.

In addition to these models, another model is trained for the original JPEG images, $I_{org}$, for comparison purposes. Furthermore, to observe the effect of standard deviation variation for random perturbation, a separate model is trained for each different standard deviation. In total, 14 different models are trained for the vehicle classification task and 4 different models for vehicle detection task.

Figure \ref{fig6.1_ThesisVehicleMainFigures} illustrates the four different input images generated from the bitstream and fed into the YOLOv7-Tiny networks. We use the following abbreviations for vehicle detection and classification based on different image types:
\begin{itemize}
    \item $V_{bp}$: Using $I_{bp}$ (Block Partition Based) images.
    \item $V_{rp}$: Using $I_{rp}$ (Random Perturbation Images).
    \item $V_{pu}$: Using $I_{pu}$ (Prediction Unit Based) images.
    \item $V_{px}$: Using $I_{px}$ (Pixel Domain) images.
\end{itemize}

To conduct these experiments, the following hardware and software configurations are employed:
\begin{itemize}
\item Computer: A computer with an Intel(R) Core(TM) i9-9900X CPU, NVIDIA GeForce GTX 1080 Ti GPU, and 48 GB RAM running a Windows 11 64-bit operating system is used.

\item Software: The reference software for the H.265/HEVC coding standard, known as HM (version 16.20), is used for both encoding and decoding purposes. The "Main profile" is used for encoding, with 4:2:0 color encoding and a quantization parameter of 32 \cite{h265.3}.

\item Compiler: The Microsoft Visual Studio 2019 (v142) platform tool-set is used to compile the reference software.

\end{itemize}

\subsection{BIT Vehicle Dataset}
We selected the BIT dataset \cite{pixel.10} for our experiments due to its widespread use in previous research and the diverse set of images it offers for vehicle classification. The BIT-Vehicle dataset, provided by the Beijing Institute of Technology, comprises 9580 vehicle images featuring six types of vehicles: sedans, sport-utility vehicles (SUVs), microbuses, trucks, buses, and minivans. The dataset exhibits varying frequencies of vehicle types. Specifically, the number of vehicles per class is as follows: 558 buses, 883 microbuses, 476 minivans, 5922 sedans, 1392 SUVs, and 822 trucks. These images were captured by road surveillance cameras and include both day and night scenes, as well as sunny days with no background noise, rain, snow, people, or other vehicle types.

To ensure a fair comparison with previous works, the dataset was divided into a training set and a validation set with a ratio of 8:2, containing 7880 and 1970 images, respectively. This ratio was also maintained for each vehicle type to ensure a balanced representation across classes. Among these images, approximately 1000 and 250 were nighttime images for training and validation, respectively.

\subsection{Reconstruction Time Comparison}
We first measure the time taken for different steps of the HEVC reconstruction process. Then, we calculate the image reconstruction time for these methods. Finally, we measure the total elapsed time, including both the reconstruction and inference time for vehicle detection and classification.

\subsubsection{Measurement of Reconstruction Steps}

The following processes are measured to determine the reconstruction time of each method.

\begin{itemize}
\item Entropy Decoding (ED): This is a common step for both methods.
\item Intra Prediction (IP): This is a common step for both methods.
\item Residual Decompression (RD): This step is skipped for $I_{rp}$.
\item Loop Filters (LF): This is a common step for both methods.
\end{itemize}

\begin{table}
{
\caption{Measurement of Reconstruction Steps.}
\begin{center}
\begin{tabular*}{\columnwidth}{l@{\extracolsep{\fill}}cccc}
\hline
\hline
Time (ms)  & Average  & Min. & Max.  \\ 
\hline
Entropy Decoding     & 6.59  & 6.00 & 8.00  \\
Intra Prediction       & 7.04  & 4.00 & 10.00 \\ 
Residual Construction       & 12.92 & 8.00 & 17.00 \\ 
Loop Filters     & 9.70  & 8.00 & 16.00 \\ 
\hline
\end{tabular*}
\end{center}
\label{table_ReconstructionSteps}}
\end{table}

\begin{table}
  \centering
  \caption{Comparison of Image Reconstruction Times.}
  \label{table_ImageGeneration}\begin{tabular*}{\columnwidth}{l@{\extracolsep{\fill}}c c c }
\hline
\hline
\textbf{Time(ms)} & $I_{rp}$ &  $I_{px}$ \\ \hline
Average  & 23.33 & 36.25 \\ \hline
Minimum  & 18.00 &  26.00 \\ \hline
Maximum  & 34.00 & 51.00 \\ \hline
\end{tabular*}
\end{table}

\subsubsection{Comparison of Image Reconstruction Time}

The elapsed time to reconstruct $I_{rp}$, denoted as $T(I_{rp})$, can be calculated using the following equation:

\begin{equation}
T(I_{rp}) = T(ED) + T(IP) + T(LF)
\end{equation}

Similarly, the elapsed time to reconstruct $I_{px}$, denoted as $T(I_{px})$, can be calculated as:

\begin{equation}
T(I_{px}) = T(ED) + T(IP) + T(RD) + T(LF)
\end{equation}

The results presented in Table \ref{table_ImageGeneration} demonstrate that the proposed method has significantly faster construction times compared to traditional full decoding. The time required to generate an image in the compressed domain using the Random Perturbation Image ($I_{rp}$) method averaged 23.33 ms, a 35.6\% reduction in time compared to the pixel domain's 36.25 ms. These results indicate that the proposed methods can substantially decrease the time required for image generation.

\subsubsection{Comparison of Total Elapsed Time}
The total time spent for both reconstruction and classification is presented in Table \ref{table_VehicleWhole}. Vehicle classification using the YOLO Convolutional Neural Network (CNN) takes approximately 2ms for both pixel and compressed domain methods, as it performs detection and classification simultaneously. The key difference between the two methods lies in the average reconstruction time. The compressed domain method ($V_{rp}$) is significantly faster, taking only 25.33ms compared to the pixel domain method ($V_{px}$) which takes 38.24ms. This demonstrates the efficiency of our compressed domain method in reducing the overall time spent on reconstruction and classification tasks.

\begin{table}
  \centering
  \caption{Comparison of the Total Time Spend for Reconstruction and Classification.}
  \label{table_VehicleWhole}\begin{tabular*}{\columnwidth}{l@{\extracolsep{\fill}}cc}
\hline
  \hline Method & Time (ms) \\
  \hline  
        $V_{rp}$ & 25.33 \\ 
        $V_{px}$ & 38.24 \\ 
          \hline
\end{tabular*}
\end{table}

\subsection{Accuracy Comparison}
\subsubsection{Metrics}
The results are evaluated using the F1-score, precision (the proportion of correct detections among all positive predictions), recall (the proportion of correct detections among all instances of the class in the dataset), average intersection of union (Avg. IoU), and mean average precision (mAP) at the IoU threshold of 0.50 (mAP@0.50).

mAP@0.50 is a metric commonly used to evaluate the performance of object detection algorithms. It is calculated as the mean of the Average Precision (AP) for each class in a dataset \eqref{eq6.1}.

\begin{equation}
    mAP = \frac{1}{n}\sum_{i=1}^{n}AP_i
\label{eq6.1}
\end{equation}

where $n$ is the number of classes in the dataset and $AP_i$ is the Average Precision for class $i$.

To calculate AP, the algorithm's predictions are first sorted by their confidence scores. Then, AP is calculated as the area under the precision-recall curve with an IoU threshold of 0.50, as shown in \eqref{eq6.2}.

\begin{equation}
    AP = \frac{\sum_{k=1}^{n}P(k) \cdot rel(k)}{\sum_{k=1}^{n}rel(k)}
\label{eq6.2}
\end{equation}

where $P(k)$ is the precision at cut-off $k$, and $rel(k)$ is a binary indicator of whether the prediction at cut-off $k$ is a true positive or not, considering an IoU threshold of 0.50.

\subsubsection{Vehicle Detection Accuracy}
The BIT Vehicle dataset is originally annotated for six different vehicle types. To measure the vehicle detection performance, the dataset has been re-annotated, combining all vehicle types into a single "vehicle" class. The models are retrained using these four different image types. The obtained vehicle detection performance is presented in Table \ref{table_VehicleDetection}.

\begin{table}
  \centering
\caption{Vehicle Detection Accuracy for Different Methods.}
\label{table_VehicleDetection}
\begin{tabular*}{\columnwidth}{l@{\extracolsep{\fill}}cc}
\hline
\hline
Method & Mean Average Precision(mAP)  \\ 
\hline
$V_{bp}$ & 98.34\%  \\ 
$V_{pu}$ & 99.89\%  \\ 
$V_{rp}$ & 99.99\% \\ 
$V_{px}$ & 99.99\% \\ 
\hline
\end{tabular*}
\end{table}

The results suggest that the Random Perturbation Image Reconstruction is able to achieve a Mean Average Precision (mAP) of 99.99\% for vehicle detection, which is comparable to the mAP of the pixel domain approach, which is also 99.99\%. The proposed compressed domain method is able to perform vehicle detection at a similar level to the pixel domain approach. 

\subsubsection{Vehicle Classification Accuracy}

\begin{table}[t!]
{
\caption{Vehicle Classification Accuracy Comparison of Random Perturbation Images Generated with Various Standard Deviations.}
\begin{center}
\resizebox{\columnwidth}{!}{%
\begin{tabular}{ c c c c c c c c c}
\hline
\hline
$V_{rp}$ $\sigma$ & Bus & Microbus & Minivan & Sedan & SUV & Truck & mAP@0.50 (\%) \\ \hline
1 & 99.46\% & 93.40\% & 93.12\% & 99.54\% & 93.98\% & 97.25\% & 96.13\% \\  
2 & 99.95\% & 95.71\% & 91.04\% & 99.61\% & 94.18\% & 97.03\% & 96.25\% \\  
3 & 99.97\% & 96.38\% & 92.59\% & 99.40\% & 93.94\% & 95.11\% & 96.23\% \\  
4 & 99.92\% & 94.14\% & 91.34\% & 99.72\% & 95.54\% & 94.56\% & 95.87\% \\
5 & 99.91\% & 94.19\% & 92.43\% & 99.64\% & 94.40\% & 97.00\% & 96.26\% \\ 
6 & 99.92\% & 95.33\% & 91.57\% & 99.65\% & 95.73\% & 96.47\% & 96.45\% \\
7& 99.99\% & 95.75\% & 93.08\% & 99.48\% & 95.04\% & 97.69\% & \textbf{96.84\%} \\
8 & 99.28\% & 95.06\% & 92.52\% & 99.67\% & 93.79\% & 96.62\% & 96.16\% \\
9 & 99.91\% & 94.34\% & 91.09\% & 99.57\% & 95.47\% & 97.73\% & 96.35\% \\
10 & 99.84\% & 94.62\% & 90.09\% & 99.52\% & 93.11\% & 98.16\% & 95.89\% \\ \hline
\end{tabular}}
\end{center}
\label{table_Accuracy_stdDev}}
\end{table}

Table \ref{table_Accuracy_stdDev} presents the vehicle classification accuracy comparison for random perturbation images generated with various standard deviations. Each row corresponds to the classification accuracy obtained using a different standard deviation value for the random perturbations. The columns represent the Average Precision (AP)  for each vehicle type, as well as the mean average precision (mAP) at an IoU threshold of 0.50.

From the results, it can be observed that for most vehicle types, the classification accuracy remains relatively high across different standard deviation values. The highest mAP is obtained when the standard deviation is set to 7, with a value of 96.84\%. This indicates that the proposed method performs well in the classification task for various standard deviations.

It is also worth noting that certain vehicle types, such as buses and sedans, consistently have higher classification accuracies than others, such as microbuses and minivans. This might be due to the distinctive features of these vehicle types, which make them easier to classify, as well as the higher frequency of sedans in the dataset. Overall, the table demonstrates the effectiveness of the proposed method for vehicle classification across a range of standard deviation values.

\begin{table*}[!ht]
{
\caption{Vehicle Classification Accuracy for Different Methods.}
\begin{center}
\begin{tabular*}{\linewidth}{l@{\extracolsep{\fill}}c c c c c c c c c c}
\hline
\hline
Implementation & Method & Domain & Bus & Microbus & Minivan & Sedan & SUV & Truck & mAP@0.50 (\%) \\ \hline
\cite{pixel.11} & YOLOv3.Tiny & Pixel & - & - & - & - & - & - & 95.05\% \\  
\cite{pixel.12} & CNN & Pixel & - & - & - & - & - & - & 93.90\% \\  
\cite{pixel.13} & YOLOv2\_Vehicle & Pixel & 98.42\% & 97.04\% & 95.02\% & 97.37\% & 93.73\% & 97.80\% & 96.56\% \\  
\cite{pixel.14} & Improved YOLOv2 & Pixel & 98.86\% & 96.63\% & 95.90\% & 98.23\% & 94.86\% & \textbf{99.30\%} & 97.30\% \\  
\cite{pixel.14} & YOLOv2 \cite{yolov2} & Pixel & 98.34\% & 95.03\% & 91.11\% & 97.42\% & 93.62\% & 98.41\% & 95.65\% \\  
\cite{pixel.14} & YOLOv3 \cite{redmon2018yolov3} & Pixel & 98.65\% & 96.98\% & 94.04\% & 97.65\% & 94.36\% & 98.17\% & 96.64\% \\  
\cite{pixel.14} & Faster R-CNN VGG16 \cite{rcnn} & Pixel & 99.05\% & 93.75\% & 91.38\% & 98.14\% & 94.75\% & 98.17\% & 95.87\% \\  
\cite{pixel.14} & SSD300 VGG16 \cite{ssd} & Pixel & 97.97\% & \textbf{97.98}\% & 90.28\% & 97.15\% & 91.25\% & 97.75\% & 93.75\% \\  
Ours & YOLOv7.Tiny $I_{org}$ & Pixel & \textbf{100.00\%} & 97.70\% & \textbf{96.40\%} & \textbf{99.76\%} & \textbf{96.49\%} & 99.23\% & \textbf{98.26\%} \\  
Ours & YOLOv7.Tiny $V_{px}$ & Pixel & \textbf{100.00\%} & 96.10\% & 96.36\% & \textbf{99.76\%} & 95.95\% & 98.74\% & 97.82\% \\  
Ours & YOLOv7.Tiny $V_{bp}$ \cite{beratoglu01} & Compressed & 83.52\% & 65.40\% & 63.51\% & 87.36\% & 66.61\% & 84.29\% & 75.11\% \\  
Ours & YOLOv7.Tiny $V_{pu}$ \cite{beratoglu02} & Compressed & 99.83\% & 92.22\% & 89.16\% & 99.51\% & 93.02\% & 98.37\% & 95.35\% \\  
Ours & YOLOv7.Tiny $V_{rp}$ $\sigma=7$ & Compressed & 99.99\% & 95.75\% & 93.08\% & 99.48\% & 95.04\% & 97.69\% & 96.84\% \\ \hline
\end{tabular*}
\end{center}
\label{table_Accuracy}}
\end{table*}

Table \ref{table_Accuracy} presents the vehicle classification accuracy for different methods in both pixel and compressed domains, comparing our proposed compressed domain method with results from the literature and pixel domain methods. The columns represent the implementation, method, domain, the Average Precision (AP) for each vehicle type, and the mean average precision (mAP) at an IoU threshold of 0.50.

In the given table, there is a distinction between the $I_{org}$ and $I_{px}$ results. The $I_{org}$ refers to the original JPEG files in the BIT dataset, while $I_{px}$ represents the encoded and re-decoded versions of the $I_{org}$ images. The purpose of this comparison is to measure the potential performance loss caused by lossy compression. As a result, the performance of $I_{org}$ is higher than that of $I_{px}$, indicating that lossy compression may have a negative impact on the classification accuracy. It is important to note that this effect is also applicable to the compressed domain experiments, such as $V_{rp}$, $V_{bp}$, and $V_{pu}$. 

For our proposed compressed domain method, the YOLOv7.Tiny $V_{rp}$ with $\sigma=7$ achieves an impressive mAP of 96.84\%, indicating that the method performs well even in the compressed domain. Although this performance is slightly lower than the pixel domain performance (97.82\%, YOLOv7.Tiny $I_{px}$), it still demonstrates the potential of compressed domain approaches for traffic surveillance applications. In comparison to the literature, the proposed compressed domain method outperforms some of the pixel domain methods, such as YOLOv3.Tiny, CNN,  YOLOv2\_Vehicle, YOLOv2, YOLOv3, SSD300 VGG16, and Faster R-CNN VGG16. This indicates that the compressed domain approach can provide a viable alternative to pixel domain methods for certain applications, especially when considering the significant speedup in the image reconstruction process.

Our compressed domain method, YOLOv7.Tiny $V_{rp}$, offers better accuracy than the other compressed domain methods YOLOv7.Tiny $V_{pu}$ and YOLOv7.Tiny $V_{bp}$, with mAP values of 95.35\% and 75.11\%, respectively. This demonstrates the effectiveness of the proposed random perturbation-based approach.

When comparing the pixel domain methods, it is evident that our YOLOv7-based models excel in this domain. Although the main focus of our research is on compressed domain methods, the YOLOv7-based models have demonstrated their superiority over other pixel domain approaches such as YOLOv3.Tiny, CNN, YOLOv2\_Vehicle, Improved YOLOv2, YOLOv2, YOLOv3, SSD300 VGG16, and Faster R-CNN VGG16.

\section{Conclusion}
\label{sec:conclusion}

In this study, we presented a novel method for reconstructing images from HEVC bitstreams, specifically designed for traffic surveillance applications. Our method replaces the calculation of residual values with random perturbations, resulting in a condensed representation of the original image while retaining the information necessary for video understanding tasks. By operating in the compressed domain, our approach significantly reduces both the computational overhead associated with full video decoding and the amount of data needed for image reconstruction, making it well-suited for real-time traffic monitoring and management applications.

Experimental results on the public BIT-Vehicle dataset demonstrated the effectiveness of our proposed method, showing a 56\% faster reconstruction process compared to the pixel domain method while maintaining high detection and classification accuracy. These findings suggest that compressed domain methods have the potential to overcome performance bottlenecks in video analytics and provide efficient solutions for traffic surveillance applications where speed and data size are critical factors.

Another significant application of the study could be the scanning of video archives. Vehicle and license plate searches can be performed on compressed videos by leveraging the method's speed advantage.

As future work, we plan to extend our method to handle both intra and inter encoded bitstreams and further explore the potential benefits of combining compressed domain attributes with pixel domain information. Combining compressed domain attributes with pixel domain information could potentially yield even better accuracy for object detection tasks. Additionally, we aim to investigate the applicability of our method to other object detection applications, such as pedestrian detection and license plate recognition. Furthermore, the data minimization aspect of our method could be particularly useful in complying with data protection regulations, such as the GDPR and CPRA, making it a valuable contribution to the development of privacy-preserving video analytics techniques.

Overall, our study contributes to the growing body of research on compressed domain methods for video understanding tasks and highlights their potential for addressing the computational and data size challenges faced in traffic surveillance and other video analytics applications.





\bibliographystyle{IEEEtran} 
\bibliography{references.bib}

\end{document}